\documentclass[letterpaper]{article} 
\usepackage{iclr2016_conference,times}
\usepackage{hyperref}
\usepackage{url}
\usepackage{makeidx}
\usepackage{graphicx}
\usepackage{url}
\usepackage{algpseudocode}
\usepackage[ruled,section]{algorithm}
\usepackage{booktabs}
\usepackage{enumitem}
\usepackage{multirow}
\usepackage{amsmath}
\usepackage{amssymb}
\usepackage{booktabs}
\usepackage{pbox}
\usepackage{xcolor}
\usepackage{color, colortbl}
\iclrfinalcopy
\title{Session-based Recommendations with \\ Recurrent Neural Networks}

\author{Bal\'azs Hidasi
    \thanks{The author spent 3 months at Telefonica Research during the research of this topic.} \\
    Gravity R\&D Inc. \\
    Budapest, Hungary\\
    \texttt{balazs.hidasi@gravityrd.com}
\And
Alexandros Karatzoglou \\
    Telefonica Research \\
    Barcelona, Spain\\
    \texttt{alexk@tid.es}
\And
Linas Baltrunas \thanks{This work was done while the author was a member of the Telefonica Research group in Barcelona, Spain}\\
     Netflix\\
    Los Gatos, CA, USA\\
    \texttt{lbaltrunas@netflix.com}
\And
Domonkos Tikk \\
    Gravity R\&D Inc. \\
    Budapest, Hungary\\
    \texttt{domonkos.tikk@gravityrd.com}
}

\begin{document}

\maketitle

\begin{abstract}
We apply recurrent neural networks (RNN) on a new domain, namely recommender systems. Real-life recommender systems often face the problem of having to base recommendations only on short session-based data (e.g. a small sportsware website) instead of long user histories (as in the case of Netflix). In this situation the frequently praised matrix factorization approaches are not accurate. This problem is usually overcome in practice by resorting to item-to-item recommendations, i.e. recommending similar items. We argue that by modeling the whole session, more accurate recommendations can be provided. We therefore propose an RNN-based approach for session-based recommendations. Our approach also considers practical aspects of the task and introduces several modifications to classic RNNs such as a ranking loss function that make it more viable for this specific problem. Experimental results on two data-sets show marked improvements over widely used approaches. 
\end{abstract}

\section{Introduction}\label{sec:intro}


Session-based recommendation is a relatively unappreciated problem in the machine learning and recommender systems community.
Many e-commerce recommender systems (particularly those of small retailers) and most of news and media sites do not typically track the user-id's of the users that visit their sites over a long period of time. While cookies and browser fingerprinting can provide some level of user recognizability, those technologies are often not reliable
enough and moreover raise privacy concerns. Even if tracking is possible, lots of users have only one or two sessions on a smaller e-commerce site, and in certain domains (e.g. classified sites) the behavior of users often shows session-based traits. Thus subsequent sessions of the same user should be handled independently. Consequently, most session-based recommendation systems deployed for e-commerce are based on relatively simple methods
that do not make use of a user profile e.g. item-to-item similarity, co-occurrence, or transition probabilities.
While effective, those methods often take only the last click or selection of the user into account ignoring the information of past clicks.

The most common methods used in recommender systems are factor models~\citep{koren2009matrix,weimer2007maximum,itals_ecml} and neighborhood methods \citep{sarwar2001item,Koren2008KDD}. Factor models work by decomposing the sparse user-item interactions matrix to a set of $d$ dimensional vectors one for each item and user in the dataset. The recommendation problem is then treated as a matrix completion/reconstruction problem whereby the latent factor vectors are then used to fill the missing entries by e.g. taking the dot product of the corresponding user--item latent factors.
Factor models are hard to apply in session-based recommendation due to the absence of a user profile. On the other hand, neighborhood methods, which rely on computing similarities between items (or users) are based on co-occurrences of items in sessions (or user profiles). Neighborhood methods have been used extensively in session-based recommendations.

The past few years have seen the tremendous success of deep neural networks in a number of tasks such as image and speech recognition \citep{Russakovsky15,hinton2012deep} where unstructured data is processed through several convolutional and standard layers of (usually rectified linear) units.
Sequential data modeling has recently also attracted a lot of attention with various flavors of RNNs being the model of choice for this type of data.
Applications of sequence modeling range from test-translation to conversation modeling to image captioning.

While RNNs have been applied to the aforementioned domains with remarkable success little attention, has been paid to the area of recommender systems.
In this work we argue that RNNs can be applied to session-based recommendation with remarkable results, we deal with the issues that arise when modeling
such sparse sequential data and also adapt the RNN models to the recommender setting by introducing a new ranking loss function suited to the task of training
these models. The session-based recommendation problem shares some similarities with some NLP-related problems in terms of modeling as long as they both deals with sequences. In the session-based recommendation we can consider the first item a user clicks when entering a web-site as the initial input of the RNN, we then would like to query the model based on this initial input for a recommendation. Each consecutive click of the user will then produce an output (a recommendation) that depends on all the previous clicks. Typically the item-set to choose from in recommenders systems can be in the tens of thousands or even hundreds of thousands. Apart from the large size of the item set, another challenge is that click-stream datasets are typically quite large thus training time and scalability are really important. As in most information retrieval and recommendation settings, we are interested in focusing the modeling power on the top-items that the user might be interested in, to this end we use ranking loss function to train the RNNs.

\section{Related work}\label{sec:related}

\subsection{Session-based recommendation}
Much of the work in the area of recommender systems has focused on models that work when a user identifier is available and a clear user profile can be built. In this setting, matrix factorization methods and neighborhood models have dominated the literature and are also employed on-line.
One of the main approaches that is employed in session-based recommendation and a natural solution to the problem of a missing user profile is
the item-to-item recommendation approach \citep{sarwar2001item,linden2003amazon} in this setting an item to item similarity matrix is precomputed
from the available session data, that is items that are often clicked together in sessions are deemed to be similar. This similarity matrix
is then simply used during the session to recommend the most similar items to the one the user has currently clicked. While simple, this
method has been proven to be effective and is widely employed. While effective, these methods are only taking  into account the last click of the user, in effect ignoring the information of the past clicks.

A somewhat different approach to session-based recommendation are Markov Decision Processes (MDPs) \citep{shani2002mdp}. MDPs are models of sequential stochastic decision problems. An MDP is defined as a four-tuple $\langle S, A, Rwd, tr \rangle$ where $S$ is the set of states, $A$ is a set of actions $Rwd$ is a reward function and $tr$ is the state-transition function. In recommender systems actions can be equated with recommendations and the simplest MPDs are essentially first order Markov chains where the next recommendation can be simply computed on the basis of the transition probability between items. The main issue with applying Markov chains in session-based recommendation is that the state space quickly becomes unmanageable when trying to include all possible sequences of user selections.

The extended version of the General Factorization Framework (GFF) \citep{gff_dmkd} is capable of using session data for recommendations. It models a session by the sum of its events. It uses two kinds of latent representations for items, one represents the item itself, the other is for representing the item as part of a session. The session is then represented as the average of the feature vectors of part-of-a-session item representation. However, this approach does not consider any ordering within the session.

\subsection{Deep Learning in Recommenders}
One of the first related methods in the neural networks literature where the use of Restricted Boltzmann Machines (RBM) for Collaborative Filtering \citep{salakhutdinov2007restricted}. In this work an RBM is used to model user-item interaction and perform recommendations. This model has been shown to be one of the best performing Collaborative Filtering models. Deep Models have been used to extract features from unstructured content such as music or images that are then used together with more conventional collaborative filtering models. In \citet{van2013deep} a convolutional deep network is used to extract feature from music files that are then used in a factor model. More recently \citet{Wang2015} introduced a more generic approach whereby a deep network is used to extract generic content-features from any types of items, these features are then incorporated in a standard collaborative filtering model to enhance the recommendation performance. This approach seems to be particularly useful in settings where there is not sufficient user-item interaction information.

\section{Recommendations with RNNs}\label{sec:rnn4rec} 
Recurrent Neural Networks have been devised to model variable-length sequence data. The main difference between RNNs and conventional feedforward deep models is the existence of an internal hidden state in the units that compose the network. Standard RNNs update their hidden state $h$ using the following update function:
\begin{equation}
\mathbf{h_t} = g(W\mathbf{x_t} + U\mathbf{h_{t-1}})
\end{equation}
Where
$g$ is a smooth and bounded function such as a logistic sigmoid function $\mathbf{x_t}$ is the input of the unit at time $t$. An RNN outputs a probability distribution over the next element of the sequence, given
its current state $\mathbf{h_t}$.

A Gated Recurrent Unit (GRU) \citep{cho2014properties} is a more elaborate model of an RNN unit that aims at dealing with the vanishing gradient problem. GRU gates essentially learn when and by how much to update the hidden state of the unit. The activation of the GRU is a linear interpolation between the previous activation and the candidate activation $\mathbf{\hat{h_t}}$:
\begin{equation}
\mathbf{h_t} = (1-\mathbf{z_t})\mathbf{h_{t-1}} + \mathbf{z_t}\mathbf{\hat{h_t}}
\end{equation}
where the update gate is given by:
\begin{equation}
\mathbf{z_t} = \sigma(W_z\mathbf{x_t} + U_z\mathbf{h_{t-1}})
\end{equation}
while the candidate activation function $\mathbf{\hat{h_t}}$ is computed in a similar manner:
\begin{equation}
\mathbf{\hat{h_t}} = \mathrm{tanh}\left({W\mathbf{x_t} + U(\mathbf{r_t} \odot \mathbf{h_{t-1}})}\right)
\end{equation}
and finaly the reset gate $\mathbf{r_t}$ is given by:
\begin{equation}
\mathbf{r_t} = \sigma(W_r\mathbf{x_t} + U_r\mathbf{h_{t-1}})
\end{equation}

\subsection{Customizing the GRU model}
We used the GRU-based RNN in our models for session-based recommendations. The input of the network is the actual state of the session while the output is the item of the next event in the session. The state of the session can either be the item of the actual event or the events in the session so far. In the former case 1-of-N encoding is used, i.e. the input vector's length equals to the number of items and only the coordinate corresponding to the active item is one, the others are zeros. The latter setting uses a weighted sum of these representations, in which events are discounted if they have occurred earlier. For the stake of stability, the input vector is then normalized. We expect this to help because it reinforces the memory effect: the reinforcement of very local ordering constraints which are not well captured by the longer memory of RNN. We also experimented with adding an additional embedding layer, but the 1-of-N encoding always performed better.

\begin{figure}[!h]
\centering
\includegraphics[width=0.3\textwidth, angle=-90]{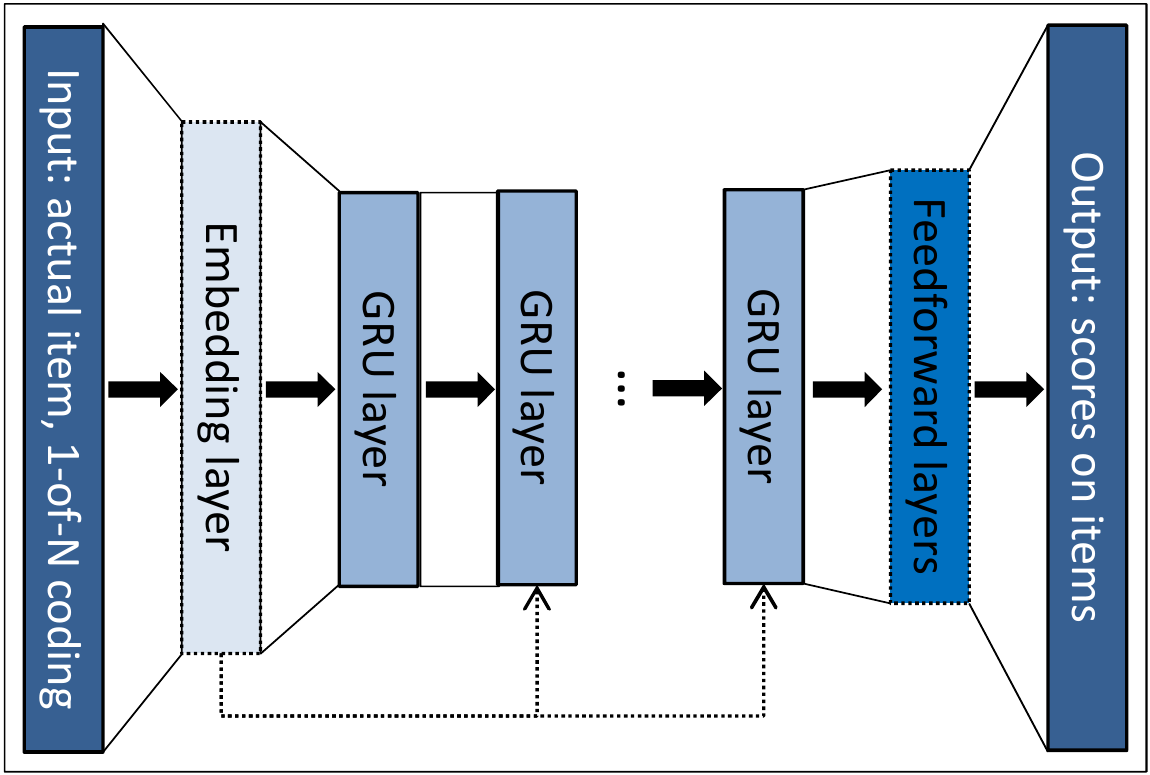}
\caption{General architecture of the network. Processing of one event of the event stream at once.}
\label{fig:arch}
\end{figure}

The core of the network is the GRU layer(s) and additional feedforward layers can be added between the last layer and the output. The output is the predicted preference of the items, i.e. the likelihood of being the next in the session for each item. When multiple GRU layers are used, the hidden state of the previous layer is the input of the next one. The input can also be optionally connected to GRU layers deeper in the network, as we found that this improves performance. See the whole architecture on Figure~\ref{fig:arch}, which  depicts the representation of a single event within a time series of events.

Since recommender systems are not the primary application area of recurrent neural networks, we modified the base network to better suit the task. We also considered practical points so that our solution could be possibly applied in a live environment.

\subsubsection{Session-parallel mini-batches}
\begin{figure}[!h]
\centering
\includegraphics[width=0.7\textwidth]{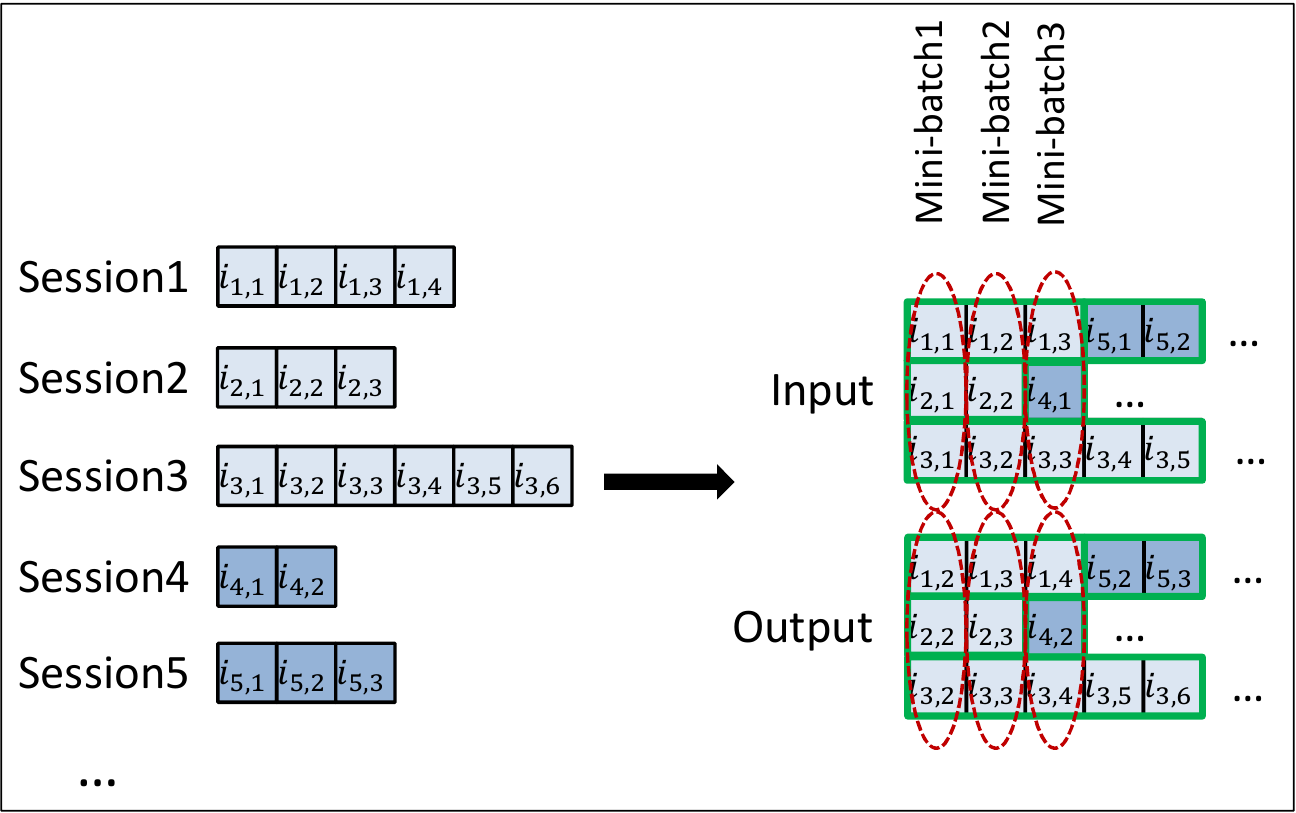}
\caption{Session-parallel mini-batch creation}
\label{fig:mini-batch}
\end{figure}

RNNs for natural language processing tasks usually use in-sequence mini-batches. For example it is common to use a sliding window over the words of sentences and put these windowed fragments next to each other to form mini-batches. This does not fit our task, because (1) the length of sessions can be very different, even more so than that of sentences: some sessions consist of only 2 events, while others may range over a few hundreds; (2) our goal is to capture how a session evolves over time, so breaking down into fragments would make no sense. Therefore we use session-parallel mini-batches. First, we create an order for the sessions. Then, we use the first event of the first $X$ sessions to form the input of the first mini-batch (the desired output is the second events of our active sessions). The second mini-batch is formed from the second events and so on. If any of the sessions end, the next available session is put in its place. Sessions are assumed to be independent, thus we reset the appropriate hidden state when this switch occurs. See Figure~\ref{fig:mini-batch} for more details.

\subsubsection{Sampling on the output}
Recommender systems are especially useful when the number of items is large. Even for a medium-sized webshop this is in the range of tens of thousands, but on larger sites it is not rare to have hundreds of thousands of items or even a few millions. Calculating a score for each item in each step would make the algorithm scale with the product of the number of items and the number of events. This would be unusable in practice. Therefore we have to sample the output and only compute the score for a small subset of the items. This also entails that only some of the weights will be updated. Besides the desired output, we need to compute scores for some negative examples and modify the weights so that the desired output is highly ranked.

The natural interpretation of an arbitrary missing event is that the user did not know about the existence of the item and thus there was no interaction. However there is a low probability that the user did know about the item and chose not to interact, because she disliked the item. The more popular the item, the more probable it is that the user knows about it, thus it is more likely that a missing event expresses dislike. Therefore we should sample items in proportion of their popularity. Instead of generating separate samples for each training example, we use the items from the other training examples of the mini-batch as negative examples. The benefit of this approach is that we can further reduce computational times by skipping the sampling. Additionally, there are also benefits on the implementation side from making the code less complex to faster matrix operations. Meanwhile, this approach is also a popularity-based sampling, because the likelihood of an item being in the other training examples of the mini-batch is proportional to its popularity.

\subsubsection{Ranking loss}
The core of recommender systems is the relevance-based ranking of items. Although the task can also be interpreted as a classification task, learning-to-rank approaches \citep{bpr,climf,gaussian_ranking} generally outperform other approaches. Ranking can be pointwise, pairwise or listwise. Pointwise ranking estimates the score or the rank of items independently of each other and the loss is defined in a way so that the rank of relevant items should be low. Pairwise ranking compares the score or the rank of pairs of a positive and a negative item and the loss enforces that the rank of the positive item should be lower than that of the negative one. Listwise ranking uses the scores and ranks of all items and compares them to the perfect ordering. As it includes sorting, it is usually computationally more expensive and thus not used often. Also, if there is only one relevant item -- as in our case -- listwise ranking can be solved via pairwise ranking.

We included several pointwise and pairwise ranking losses into our solution. We found that pointwise ranking was unstable with this network (see Section~\ref{sec:experiments} for more comments). Pairwise ranking losses on the other hand performed well. We use the following two.
\begin{itemize}
\item \textbf{BPR}: Bayesian Personalized Ranking \citep{bpr} is a matrix factorization method that uses pairwise ranking loss. It compares the score of a positive and a sampled negative item. Here we compare the score of the positive item with several sampled items and use their average as the loss. The loss at a given point in one session is defined as: $L_s=-\frac{1}{N_S}\cdot\sum_{j=1}^{N_S}{\mathrm{log}\left(\sigma\left(\hat{r}_{s,i}-\hat{r}_{s,j}\right)\right)}$, where $N_S$ is the sample size, $\hat{r}_{s,k}$ is the score on item $k$ at the given point of the session, $i$ is the desired item (next item in the session) and $j$ are the negative samples.
\item \textbf{TOP1}: This ranking loss was devised by us for this task. It is the regularized approximation of the relative rank of the relevant item. The relative rank of the relevant item is given by $\frac{1}{N_S}\cdot\sum_{j=1}^{N_S}{I\{\hat{r}_{s,j}>\hat{r}_{s,i}\}}$. We approximate $I\{\cdot\}$ with a sigmoid. Optimizing for this would modify parameters so that the score for $i$ would be high. However this is unstable as certain positive items also act as negative examples and thus scores tend to become increasingly higher. To avoid this, we want to force the scores of the negative examples to be around zero. This is a natural expectation towards the scores of negative items. Thus we added a regularization term to the loss. It is important that this term is in the same range as the relative rank and acts similarly to it. The final loss function is as follows: $L_s=\frac{1}{N_S}\cdot\sum_{j=1}^{N_S}{\sigma\left(\hat{r}_{s,j}-\hat{r}_{s,i}\right)+\sigma\left(\hat{r}_{s,j}^2\right)}$
\end{itemize}

\section{Experiments}\label{sec:experiments}
We evaluate the proposed recursive neural network against popular baselines on two datasets.

The first dataset is that of RecSys Challenge 2015\footnote{http://2015.recsyschallenge.com/}. This dataset contains click-streams of an e-commerce site that sometimes end in purchase events. We work with the training set of the challenge and keep only the click events. We filter out sessions of length 1. The network is trained on $\sim6$ months of data, containing 7,966,257 sessions of 31,637,239 clicks on 37,483 items. We use the sessions of the subsequent day for testing. Each session is assigned to either the training or the test set, we do not split the data mid-session. Because of the nature of collaborative filtering methods, we filter out clicks from the test set where the item clicked is not in the train set. Sessions of length one are also removed from the test set. After the preprocessing we are left with 15,324 sessions of 71,222 events for the test set. This dataset will be referred to as RSC15.

The second dataset is collected from a Youtube-like OTT video service platform. Events of watching a video for at least a certain amount of time were collected. Only certain regions were subject to this collection that lasted for somewhat shorter than 2 months. During this time item-to-item recommendations were provided after each video at the left side of the screen. These were provided by a selection of different algorithms and influenced the behavior of the users. Preprocessing steps are similar to that of the other dataset with the addition of filtering out very long sessions as they were probably generated by bots. The training data consists of all but the last day of the aforementioned period and has $\sim3$ million sessions of $\sim13$ million watch events on $330$ thousand videos. The test set contains the sessions of the last day of the collection period and has $\sim37$ thousand sessions with $\sim180$ thousand watch events. This dataset will be referred to as VIDEO.

The evaluation is done by providing the events of a session one-by-one and checking the rank of the item of the next event. The hidden state of the GRU is reset to zero after a session finishes. Items are ordered in descending order by their score and their position in this list is their rank. With RSC15, all of the 37,483 items of the train set were ranked. However, this would have been impractical with VIDEO, due to the large number of items. There we ranked the desired item against the most popular 30,000 items. This has negligible effect on the evaluations as rarely visited items often get low scores. Also, popularity based pre-filtering is common in practical recommender systems.

As recommender systems can only recommend a few items at once, the actual item a user might pick should be amongst the first few items of the list. Therefore, our primary evaluation metric is recall@20 that is the proportion of cases having the desired item amongst the top-20 items in all test cases. Recall does not consider the actual rank of the item as long as it is amongst the top-N. This models certain practical scenarios well where there is no highlighting of recommendations and the absolute order does not matter. Recall also usually correlates well with important online KPIs, such as click-through rate (CTR)\citep{Liu:2012EBR,itals_ecml}. The second metric used in the experiments is MRR@20 (Mean Reciprocal Rank). That is the average of reciprocal ranks of the desired items. The reciprocal rank is set to zero if the rank is above 20. MRR takes into account the rank of the item, which is important in cases where the order of recommendations matter (e.g. the lower ranked items are only visible after scrolling).

\subsection{Baselines}
We compare the proposed network to a set of commonly used baselines.
\begin{itemize}[noitemsep,nolistsep]
\item \textbf{POP}: Popularity predictor that always recommends the most popular items of the training set. Despite its simplicity it is often a strong baseline in certain domains.
\item \textbf{S-POP}: This baseline recommends the most popular items of the current session. The recommendation list changes during the session as items gain more events. Ties are broken up using global popularity values. This baseline is strong in domains with high repetitiveness.
\item \textbf{Item-KNN}: Items similar to the actual item are recommended by this baseline and similarity is defined as the cosine similarity between the vector of their sessions, i.e. it is the number of co-occurrences of two items in sessions divided by the square root of the product of the numbers of sessions in which the individual items are occurred. Regularization is also included to avoid coincidental high similarities of rarely visited items. This baseline is one of the most common item-to-item solutions in practical systems, that provides recommendations in the ``others who viewed this item also viewed these ones'' setting. Despite of its simplicity it is usually a strong baseline \citep{linden2003amazon,JamesRecsys10}.
\item \textbf{BPR-MF}: BPR-MF \citep{bpr} is one of the commonly used matrix factorization methods. It optimizes for a pairwise ranking objective function (see Section~\ref{sec:rnn4rec}) via SGD. Matrix factorization cannot be applied directly to session-based recommendations, because the new sessions do not have feature vectors precomputed. However we can overcome this by using the average of item feature vectors of the items that had occurred in the session so far as the user feature vector. In other words we average the similarities of the feature vectors between a recommendable item and the items of the session so far.
\end{itemize}

\begin{table}
\centering
\caption{Recall@20 and MRR@20 using the baseline methods}\label{tab:baselines}
\medskip
{\small
\begin{tabular}{lcccc}
\toprule
\multirow{2}{*}{\textbf{Baseline}} & \multicolumn{2}{c}{\textbf{RSC15}} & \multicolumn{2}{c}{\textbf{VIDEO}} \\
& Recall@20 & MRR@20 & Recall@20 & MRR@20 \\
\midrule
POP & 0.0050 & 0.0012 & 0.0499 & 0.0117 \\
S-POP & 0.2672 & 0.1775 & 0.1301 & 0.0863 \\
Item-KNN & 0.5065 & 0.2048 & 0.5508 & 0.3381 \\
BPR-MF & 0.2574 & 0.0618 & 0.0692 & 0.0374 \\
\bottomrule
\end{tabular}}
\end{table}

Table~\ref{tab:baselines} shows the results for the baselines. The item-KNN approach clearly dominates the other methods.

\subsection{Parameter \& structure optimization}
We optimized the hyperparameters by running 100 experiments at randomly selected points of the parameter space for each dataset and loss function. The best parametrization was further tuned by individually optimizing each parameter. The number of hidden units was set to 100 in all cases. The best performing parameters were then used with hidden layers of different sizes. The optimization was done on a separate validation set. Then the networks were retrained on the training plus the validation set and evaluated on the final test set.

The best performing parametrizations are summarized in table~\ref{tab:params}. Weight matrices were initialized by random numbers drawn uniformly from $[-x,x]$ where $x$ depends on the number of rows and columns of the matrix. We experimented with both rmsprop \citep{rmsprop} and adagrad \citep{adagrad}. We found adagrad to give better results.

\begin{table}
\centering
\caption{Best parametrizations for datasets/loss functions}\label{tab:params}
\medskip
{\small
\begin{tabular}{llcccc}
\toprule
\textbf{Dataset} & \textbf{Loss} & \textbf{Mini-batch} & \textbf{Dropout} & \textbf{Learning rate} & \textbf{Momentum} \\
\midrule
RSC15 & TOP1 & 50 & 0.5 & 0.01 & 0 \\
RSC15 & BPR & 50 & 0.2 & 0.05 & 0.2 \\
RSC15 & Cross-entropy & 500 & 0 & 0.01 & 0 \\
VIDEO & TOP1 & 50 & 0.4 & 0.05 & 0 \\
VIDEO & BPR & 50 & 0.3 & 0.1 & 0 \\
VIDEO & Cross-entropy & 200 & 0.1 & 0.05 & 0.3 \\
\bottomrule
\end{tabular}}
\end{table}

We briefly experimented with other units than GRU. We found both the classic RNN unit and LSTM to perform worse.

We tried out several loss functions. Pointwise ranking based losses, such as cross-entropy and MRR optimization (as in \citet{gaussian_ranking}) were usually unstable, even with regularization. For example cross-entropy yielded only 10 and 6 numerically stable networks of the 100 random runs for RSC15 and VIDEO respectively. We assume that this is due to independently trying to achieve high scores for the desired items and the negative push is small for the negative samples. On the other hand pairwise ranking-based losses performed well. We found the ones introduced in Section~\ref{sec:rnn4rec} (BPR and TOP1) to perform the best.

Several architectures were examined and a single layer of GRU units was found to be the best performer. Adding addition layers always resulted in worst performance w.r.t. both training loss and recall and MRR measured on the test set. We assume that this is due to the generally short lifespan of the sessions not requiring multiple time scales of different resolutions to be properly represented. However the exact reason of this is unknown as of yet and requires further research. Using embedding of the items gave slightly worse results, therefore we kept the 1-of-N encoding. Also, putting all previous events of the session on the input instead of the preceding one did not result in additional accuracy gain; which is not surprising as GRU -- like LSTM -- has both long and short term memory. Adding additional feed-forward layers after the GRU layer did not help either. However increasing the size of the GRU layer improved the performance. We also found that it is beneficial to use tanh as the activation function of the output layer.

\subsection{Results}
Table~\ref{tab:results} shows the results of the best performing networks. Cross-entropy for the VIDEO data with 1000 hidden units was numerically unstable and thus we present no results for that scenario. The results are compared to the best baseline (item-KNN). We show results with 100 and 1000 hidden units. The running time depends on the parameters and the dataset. Generally speaking the difference in runtime between the smaller and the larger variant is not too high on a GeForce GTX Titan X GPU and the training of the network can be done in a few hours\footnote{Using Theano with fixes for the subtensor operators on GPU.}. On CPU, the smaller network can be trained in a practically acceptable timeframe. Frequent retraining is often desirable for recommender systems, because new users and items are introduced frequently.

The GRU-based approach has substantial gain over the item-KNN in both evaluation metrics on both datasets, even if the number of units is 100\footnote{Except for using the BPR loss on the VIDEO data and evaluating for MRR.}. Increasing the number of units further improves the results for pairwise losses, but the accuracy decreases for cross-entropy. Even though cross-entropy gives better results with 100 hidden units, the pairwise loss variants surpass these results as the number of units increase. Although, increasing the number of units increases the training times, we found that it was not too expensive to move from 100 units to 1000 on GPU. Also, the cross-entropy based loss was found to be numerically unstable as the result of the network individually trying to increase the score for the target items, while the negative push is relatively small for the other items. Therefore we suggest using any of the two pairwise losses. The TOP1 loss performs slightly better on these two datasets, resulting in $\sim20-30\%$ accuracy gain over the best performing baseline.

\begin{table}
\centering
\caption{Recall@20 and MRR@20 for different types of a single layer of GRU, compared to the best baseline (item-KNN). Best results per dataset are highlighted.}\label{tab:results}
\medskip
{\small
\begin{tabular}{lcccc}
\toprule
\multirow{2}{*}{\textbf{Loss / \#Units}} & \multicolumn{2}{c}{\textbf{RSC15}} & \multicolumn{2}{c}{\textbf{VIDEO}} \\
& Recall@20 & MRR@20 & Recall@20 & MRR@20 \\
\midrule
TOP1 \ 100 & 0.5853 (+15.55\%) & 0.2305 (+12.58\%) & 0.6141 (+11.50\%) & 0.3511 (+3.84\%)\\
BPR \ 100 & 0.6069 (+19.82\%) & 0.2407 (+17.54\%) & 0.5999 (+8.92\%) & 0.3260 (-3.56\%)\\
Cross-entropy \ 100 & 0.6074 (+19.91\%) & 0.2430 (+18.65\%) & 0.6372 (+15.69\%) & 0.3720 (+10.04\%)\\
TOP1 \ 1000 & 0.6206 (+22.53\%) & \textbf{0.2693 (+31.49\%)} & \textbf{0.6624 (+20.27\%)} & \textbf{0.3891 (+15.08\%)}\\
BPR \ 1000 & \textbf{0.6322 (+24.82\%)} & 0.2467 (+20.47\%) & 0.6311 (+14.58\%) & 0.3136 (-7.23\%)\\
Cross-entropy \ 1000 & 0.5777 (+14.06\%) & 0.2153 (+5.16\%) & -- & -- \\
\bottomrule
\end{tabular}}
\end{table}

\section{Conclusion \& Future work}\label{conclusion}
In this paper we applied a kind of modern recurrent neural network (GRU) to new application domain: recommender systems. We chose the task of session based recommendations, because it is a practically important area, but not well researched. We modified the basic GRU in order to fit the task better by introducing session-parallel mini-batches, mini-batch based output sampling and ranking loss function. We showed that our method can significantly outperform popular baselines that are used for this task. We think that our work can be the basis of both deep learning applications in recommender systems and session based recommendations in general.

Our immediate future work will focus on the more thorough examination of the proposed network. We also plan to train the network on automatically extracted item representation that is built on content of the item itself (e.g. thumbnail, video, text) instead of the current input.

\subsubsection*{Acknowledgments}
The work leading to these results has received funding from the European Union's Seventh Framework Programme (FP7/2007-2013) under CrowdRec Grant Agreement n$^\circ$ 610594.

\bibliography{citations}
\bibliographystyle{iclr2016_conference}

\end{document}